\documentclass[10pt]{article} % For LaTeX2e
\usepackage[preprint]{tmlr}
\usepackage{amsmath,amsfonts,bm}

\def\eqref#1{equation~\ref{#1}}
\def\1{\bm{1}}

\DeclareMathAlphabet{\mathsfit}{\encodingdefault}{\sfdefault}{m}{sl}
\SetMathAlphabet{\mathsfit}{bold}{\encodingdefault}{\sfdefault}{bx}{n}

\usepackage{hyperref}
\usepackage{url}
\usepackage{graphicx}
\usepackage{booktabs}
\usepackage{amsmath,amsfonts,amssymb}
\usepackage{cleveref}

\newcommand{\gray}[1]{\textcolor{gray}{#1}}

\title{Discovering Meaningful Units with Visually Grounded \\Semantics from Image Captions}

\author{\name Melika Behjati\thanks{Work done while at Idiap Research Institute and EPFL.} \email melikabehjati@gmail.com \\
      \addr 
      \AND
      \name James Henderson \email james.henderson@idiap.ch \\
      \addr Idiap Research Institute \\ Martigny, Switzerland
      }

\def\month{MM}  % Insert correct month for camera-ready version
\def\year{YYYY} % Insert correct year for camera-ready version
\def\openreview{\url{https://openreview.net/forum?id=XXXX}} % Insert correct link to OpenReview for camera-ready version

\begin{document}

\maketitle

\begin{abstract}
Fine-grained knowledge is crucial for vision-language models to obtain a better understanding of the real world. 
While there has been work trying to acquire this kind of knowledge in the space of vision and language, it has mostly focused on aligning the image patches with the tokens on the language side. 
However, image patches do not have any meaning to the human eye, and individual tokens do not necessarily carry groundable information in the image.
% neither do the single-vector representation. 
It is groups of tokens which describe different aspects of the scene. In this work, we propose a model which groups the caption tokens as part of its architecture in order to capture a fine-grained representation of the language. 
% We align these representations with the object representations that a trained visual object discovery module outputs. 
We expect our representations to be at the level of objects present in the image, and therefore align our representations with the output of an image encoder trained to discover objects. 
We show that by learning to group the tokens, the vision-language model has a better fine-grained understanding of vision and language. In addition, the token groups that our model discovers are highly similar to groundable phrases in text, both qualitatively and quantitatively. 
\end{abstract}

\section{Introduction}
%TODO: add more prior work 
% Fine-grained knowledge is currently a challenge for vision language models and is crucial for models of language to obtain a better understanding of the real world.  While there has been work trying to acquire this kind of knowledge in the space of vision and language, most of them have focused on aligning the image patches with the tokens on the language side. 
% However, image patches do not have any meaning to the human eye and individual tokens often do not carry information groundable in the image.  Minimally, it is groups of tokens in the text which describe objects in the image.  For this reason, some research has investigated object discovery on the image side to get better correspondence with groundable phrases on the language side.  In this work, we investigate the unsupervised discovery of groundable phrases on the language side to get better correspondence with objects on the vision side.  We hypothesise that finding these meaningful units in language representations will improve the fine-grained understanding of image-caption semantic relationships. 

Vision-language models have been shown to be less effective at capturing fine-grained information (e.g., understanding relationships and recognizing verbs) about the images described by the captions \citep{yuksekgonul2022and,bugliarello-etal-2023-measuring,kamath-etal-2023-text, dumpala2024sugarcrepe++, pearson2025evaluating}. This information is crucial for the models to obtain a better understanding of the real world. While there has been work trying to acquire this kind of knowledge in the space of vision and language, it has mostly focused on aligning the image patches with the input tokens on the language side \citep{yao2021filip,wang2022multi,zeng2021multi,mukhoti2023open, zhang2024longclip}. 
However, image patches do not have any meaning to the human eye, and individual tokens often do not carry information groundable in the image, neither do the single-vector representations.  Minimally, it is groups of image patches which represent objects and the group of tokens in the text that refer to those objects. 
% we might talk about phrase grounding here
% Phrase grounding is the task which tries to directly address this problem in a supervised or weakly supervised setup, where given the phrases the model is expected to find the boundary boxes over the objects that they refer to. Recently, \citet{liu2023dq} suggested that learning the phrase boundaries as well as the object boundaries helps the model to gain a better understanding of the data and therefore, the downstream tasks. 
For this reason, there has been an active line of research in vision investigating the unsupervised discovery of objects by learning to assign image patches to their representative object slots \citep{locatello2020object,wu2024structured_world_modeling,didolkar2025ctrl}. %discovering objects is an important step (building block) towards human-level understanding. 
\citet{xu2022groupvit} integrated an object discovery module into their vision-language model to learn the object entities. They showed that representing the image at the level of its constituent objects improves the performance of their model in downstream tasks.   
%They showed that by learning the objects as part of the model, their model was able to achieve better zero-shot classification performance as well as reasonalble zero-shot semantic segmentation quality.     
% For this reason, some research has investigated object discovery on the image side to get better correspondence with groundable phrases on the language side \citep{xu2022groupvit}.  
In this paper, we investigate the unsupervised discovery of groundable phrases on the language side to get better correspondence with objects on the vision side.  We hypothesize that finding these meaningful units in language representations will improve the fine-grained understanding of image-caption semantic relationships.
As far as we are aware, we are the first to investigate this possibility.
%\melika{talk about the novelty?}

\begin{figure*}[t]
    \centering
    \includegraphics[width=0.8\textwidth]{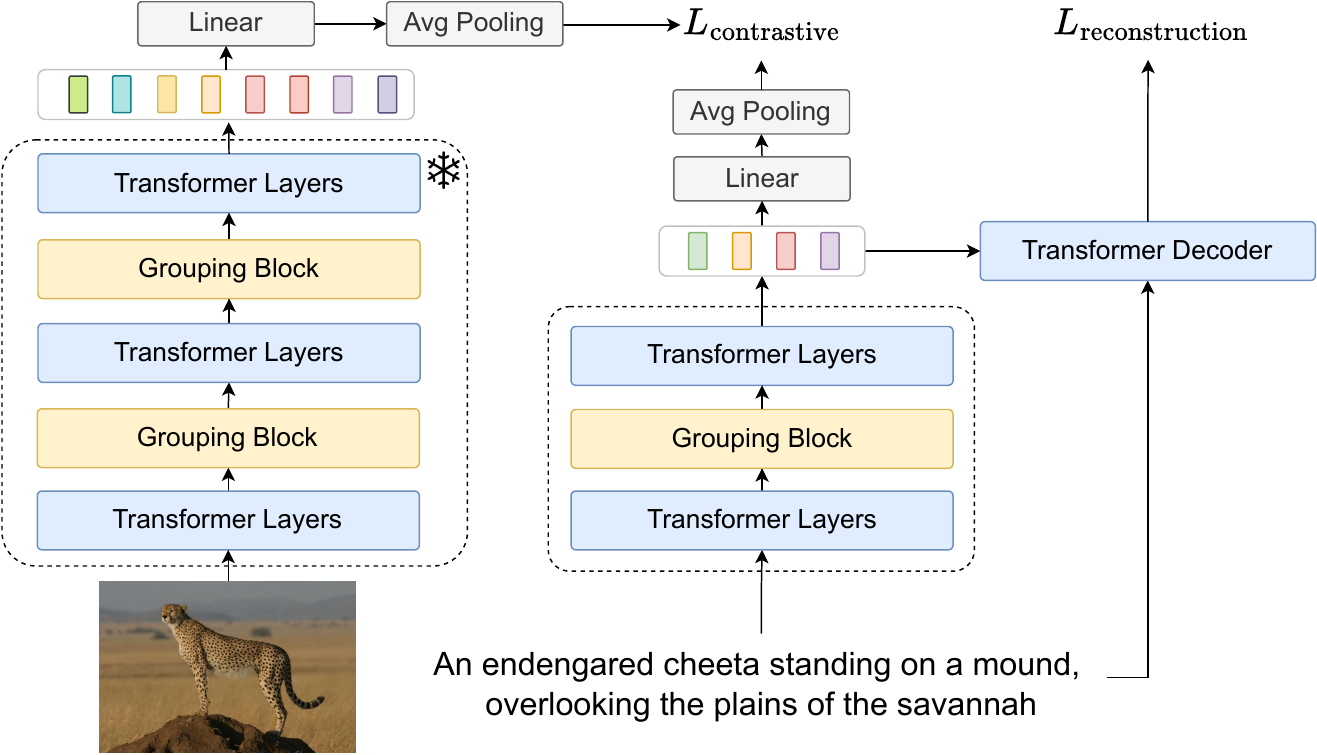}
    \caption{Overview of the model. We freeze the image encoder and only train the text encoder, decoder, and the linear projection heads.  
    The image passes through Transformer layers followed by the grouping blocks. 
    The output of the image encoder is a set of groups which are approximately representing the objects. The caption also passes through the same set of blocks and the output of the text encoder is a set of groups representing units in language. The two modalities interact via a contrastive loss. There is also a reconstruction loss where the decoder decodes the text groups into the original input. 
    }
    \label{fig:model}
\end{figure*}

We base our model on the model of visual object discovery using image caption pairs proposed by \citet{xu2022groupvit}.  We freeze the image side of the model, and introduce analogous deep learning mechanisms to discover \textit{objects}\footnote{We use the terms \textit{objects}, \textit{groups} and \textit{units} interchangeably.} on the language side (see \Cref{fig:model}). %can add explanations here 
We investigate two types of losses, one which promotes the correspondence between representations of the language side and representations on the vision side, and one which promotes the ability to reconstruct the text from the language representations.  We find that training with both these losses leads to better fine-grained understanding of the image-text relationship, and discovers units which are highly similar to groundable phrases in text, both qualitatively and quantitatively.  Further analysis finds that optimizing the image-text correspondence alone does not lead to the discovery of meaningful units on the language side, and while this model does learn a good fine-grained understanding of the image-text relationship, it does not represent the semantics of objects as well as the model which does represent groundable phrases.  We also find that optimizing the reconstruction loss alone does lead to the discovery of meaningful units on the language side, but they have a slightly worse similarity to groundable phrases than the model which includes grounding information, and do not capture image-text relationships.

% fine-grained knowledge is important.
% fine-grained representation , representing at the level of entities helps 
% Q: Can we represent language in a more fine-grained fashion? 
% Q: If yes, would that be helpful? 

% Provisional story:
% \begin{itemize}
%     \item Possible intuition: different tasks need different levels of abstraction \melika{is it a good intuition?}
%     \item Purpose: represent language at a higher level of abstraction; i.e., from subwords to phrases; which have visually grounded semantics, \melika{what are they useful for?}  
%     \item Method: a model which learns groupings as part of its architecture 
% \end{itemize}                         
% Claims
% \begin{itemize}
%     \item a more interpretable representation of language which is aligned with the semantic space of the images %interpretable in terms of attention maps
%     \item improve the language understanding of the model by discovering meaningful units % validated by svo & foil
%     \item fine-grained information about the objects helps the model to learn better representations %how is this evaluated?
    
% \end{itemize}
% entity discovery on the language side of a VL model helps it to have a better understanding of language
Our contributions are as follows, 
\begin{itemize}
    \item We develop a novel model to discover meaningful units from the image captions in the vision language setup (\Cref{sec:model}).
    \item We show that our model has better fine-grained vision and language understanding compared to a single-vector representation of text, under two different benchmarks (\Cref{sec:probe}).
    \item We show that the segments that our model discovers are meaningful both qualitatively, and in terms of accordance with human-annotated groundable phrases (\Cref{sec:attn}).
    % \item We assess the 
\end{itemize}
% (I) We develop a novel model to discover meaningful units from the image captions in the vision language setup (\Cref{sec:model}). (II) We improve the fine-grained vision and language understanding of our model compared to a single-vector representation of text, under two different benchmarks (\Cref{sec:probe}). (III)  We show that the segments that our model discovers are meaningful both qualitatively, and in terms of accordance with groundable phrases (\Cref{sec:attn}).
% \begin{itemize}
%     \item We develop a novel model to discover meaningful units from the image captions in the vision language setup (\Cref{sec:model}).
%     \item We improve the fine-grained vision and language understanding of our model compared to a single-vector representation of text, under two different benchmarks (\Cref{sec:probe}). 
%     \item We show that the segments that our model discovers are meaningful both qualitatively, and in terms of accordance with groundable phrases (\Cref{sec:attn}).
%     % \begin{itemize}
%     %     \item meaningful segmentation in term of accordance with groundable phrases + emergence of contiguous segments in the Attention maps 
%     %     \item better representation of language in the VL setup 
%     % \end{itemize}
% \end{itemize}

\section{Method}

%be consistent with the notations and also the terms used for subwords/tokens and groups/segments/
% \subsection{Problem} i.e., 
To facilitate learning the fine-grained semantics of image-text relationships, we propose a model for learning text representations whose granularity matches the granularity of objects in the image, meaning that it is neither as coarse-grained as having a single vector for embedding the entire text\footnote{This is the common way of representing text in dual-stream vision-language models like CLIP \citep{radford2021learning}.} nor as fine-grained as having a different vector for every token.
Given a dataset of image-caption pairs, $\mathrm{D}=\{(I_i,T_i)\}_{i=1,\dots,N}$, we want to learn a representation of each caption in the form of groups of tokens which are aligned with the semantic space of objects in its image. To do so, we freeze the image encoder which has been trained to output the objects in the image and only train the text encoder and the projection heads. %Given that the image encoder is outputting the objects in the image, we want to address the question: can we analogously find groups of tokens on the language side?  
The input representation of language is at the level of subwords, so we aim to find a more abstract representation which would approximately represent groundable phrases, for example as defined by the human annotators of \citet{plummer2015flickr30k}. More specifically, let $T_i=[t_{i1},\dots,t_{iM}]$, where $t_{ij}$ is a subword of $T_i$ and $M$ is the total number of subwords in $T_i$. We would like to group the subword tokens $t_{ij}$s into non-overlapping groups $T_i=\{g^T_{i1},\dots,g^T_{iK}\}$ where $K<M$ and every $t_{ij}$ is assigned to a group. This would lead to a more compact abstract representation of $T_i$.  In this work we treat the number of groups $K$ as a fixed hyperparameter similar to \citep{xu2022groupvit, seitzer2023bridging, singh2022illiterate}, leaving the identification of the number of groundable phrases in a given example to future work.

% Previous work has shown that representing images at a higher level \citep{xu2022groupvit} would lead to better performance in downstream tasks. Therefore, by having a model which is able to output meaningful groups (objects) of the image, we want to learn the groups on the text side. 

\subsection{Model}
\label{sec:model}

We illustrate an overview of our model in \Cref{fig:model}  
% The image is processed by interleaved layers of transformers and grouping blocks, resulting in a set of learned groups. The text input is also processed into a set of learned semantic groups. We then apply our two loss functions to the learned groups. 
and describe each of its components in the following sections.

\subsubsection{Text Encoder: Text Group Transformer}
% \begin{enumerate}
%     \item token embedding+pos embedding append group tokens 
%     \item Transformer encoder layer 
%     \item grouping block
%     \item Transformer encoder layers
%     \item output: a set of text groups $g^T_{ij}$
% \end{enumerate}
We design our text encoder to learn semantic units of language at the level of objects in the image, by grouping the caption tokens. The key idea is to have shared learnable group vectors which can bind to different tokens of input \citep{xu2022groupvit}. At each stage, the groups carry the information from the previous layer to the next layer.  To initiate the binding, the groups are appended to the input tokens they need to bind, and they all interact via several Transformer encoder layers to allow the groups and tokens to exchange information. Then, by performing a top-down attention mechanism shown as the Grouping block, the groups bind to different parts of the input. 

%TODO: the notations should stay consistent! 
More specifically, we first embed the input tokens and add learned positional encodings to them. Then, we append the learnable group vectors, $[g^T_{ik}]_{k=1 \dots K}$, to these embedded inputs, $[t_{ij}]_{j=1\dots M}$, and pass the resulting vectors through some Transformer encoder layers, allowing them to interact with each other. We denote the encoded tokens and groups as $\hat{t}_{ij}$ and $\hat{g}_{ik}$.  Then the grouping happens in a grouping block. In this block, the groups act as the queries and the encoded inputs as keys and values through a top-down attention mechanism.  As with standard attention, the raw attention scores are computed as
% \vspace{-1ex}
\begin{equation}
    A^{\text{raw}}_{kj} = \frac{Q(\hat{g}^T_{ik})K^\intercal(\hat{t}_{ij})}{\sqrt{d}}
% \vspace{-1ex}
\end{equation}
where $d$ is the dimension of the model and $Q$ and $K$ are linear query and key projections.  
In order to have discrete assignments of inputs to the groups, GroupViT actually performs a hard assignment over $A^{\text{raw}}$ by utilizing Gumble softmax \citep{jang2016categorical,maddison2016concrete}. 
Namely, 
% \vspace{-1ex}
\begin{equation}
    A'=\text{GumbleSoftmax}(A^{\text{raw}}).
    %=\frac{e^{A^{raw}_{i}}}{\sum_k e^{A^{raw}_{kj}}}.
% \vspace{-1ex}
\end{equation}
In top-down attention, instead of normalizing over the keys in the softmax function, the $A'$ weights are first normalized over the queries, which are the groups.
%$A'_{ij} = \frac{e^{A^{raw}_{ij}}}{\sum_i e^{A^{raw}_{ij}}}$
This will make the groups compete for representing different inputs \citep{locatello2020object} and has been shown to be the most important component in discovering objects \citep{wu2023inverted_attention}. 
After the normalization, the hard assignment happens and the gradient is backpropagated with the straight through trick \citep{van2017neural}, that is:
% \vspace{-1ex}
\begin{equation}
    A = \text{one-hot}(\text{argmax}_{\text{groups}}(A')) - \text{sg}(A') + A'
% \vspace{-0.5ex}
\end{equation}
where sg is the stop gradient operator. 
Finally, the group vectors get updated as  
% \vspace{-1ex}
\begin{equation}
    \bar{g}^T_{ik} = \hat{g}^T_{ik} + W(\sum_j \frac{A_{kj}}{\sum_j A_{kj}} V(t_{ij}))
% \vspace{-1ex}
\end{equation}
where $V$ and $W$ are the linear projections for values and outputs respectively. 
% \jamie{This equation can't be right.}

After the grouping block, the updated group vectors serve as inputs to subsequent Transformer encoder layers. 
%It is possible to have more than one grouping block, as done on the image side, by stacking transformer encoder layers and the grouping block. 
%After the last stage of grouping, 
% The resulting groups pass through some Transformer layers. 
Finally, these refined groups represent the fine-grained semantics of the text in our model. %these groups are what we use as our representation of the fine-grained semantics of the text.   

\subsubsection{Image Encoder}

We use the image encoder of \citet{xu2022groupvit}, which follows the same architecture as the text encoder, but with two stacked levels of transformer encoder layers and grouping blocks. As its input, the images are first divided into patches and then linearly projected. The encoder then extracts the set of image groups denoted as $\{\bar{g}^I_{ik}\}$.  Due to the computational cost, we freeze the image encoder and assume that the image groups are representing objects in the image. 
% The pretrained image encoder has two levels of grouping blocks, with ??? and ??? groups respectively.

\subsection{Training Objectives}

Our model is trained with two different losses, i.e., a contrastive loss and a reconstruction loss, which we will explain in the following. The two losses are combined with a hyperparameter $\lambda$ which controls the ratio between the two terms. 
% \vspace{-1ex}
\begin{equation}
    L_{\text{total}} = L_{\text{contrastive}}+\lambda L_{\text{reconstruction}}
%\vspace{-1ex}
\end{equation}

\subsubsection{Contrastive Loss}

The image and text modalities interact via a contrastive loss. First, the final groups for each modality are mapped into a common space with a Linear projector ($\Phi^T$), i.e., $z^T_{ij}=\Phi^T(\bar{g}^T_{ij})$. Then, we average pool over them to obtain the global features for each modality ($\hat{z}^T_i$). We compute the InfoNCE loss \citep{oord2018infonce} for every modality separately. Given a batch size of $B$ and a similarity function (sim), the infoNCE loss for the image to text is  
% \vspace{-1ex}
\begin{equation}
    L_\text{I-T}=-\frac{1}{B} \sum_{i=1}^B \text{log} \frac{e^{\text{sim}(\hat{z}^T_i,\hat{z}^I_i)/\tau}}{\sum_{j=1}^B e^{\text{sim}(\hat{z}^T_j,\hat{z}^I_i)/\tau} },
% \vspace{-1ex}
\end{equation}
and respectively for the text to image is
% \vspace{-1ex}
\begin{equation}
    L_\text{T-I}=-\frac{1}{B} \sum_{i=1}^B \text{log} \frac{e^{\text{sim}(\hat{z}^T_i,\hat{z}^I_i)/\tau}}{\sum_{j=1}^B e^{\text{sim}(\hat{z}^T_i,\hat{z}^I_j)/\tau}}.
% \vspace{-1ex}
\end{equation}
The final contrastive loss is calculated by averaging the two losses,
% \vspace{-1ex}
\begin{equation}
    L_\text{contrastive}= \frac{1}{2}(L_\text{I-T}+L_\text{T-I}).
% \vspace{-1ex}
\end{equation}
% \jamie{Need to define $\text{sim}(\hat{z}^T_j,\hat{z}^I_i)$.}
As for the similarity function sim(a,b), we consider the cosine similarity between the vectors.
% , i.e., 
% \begin{equation}
%     \text{sim}(a,b)=\frac{a.b}{|a||b|}
% \end{equation}

\subsubsection{Reconstruction Loss} % - encourages segmentation, 

In order to encourage the model to group the tokens into meaningful units, we incorporate a reconstruction loss from a text decoder. This loss encourages the model to assign tokens to different groups in order to spread information about the text across multiple vectors, and thus make better use of the available vectors.  
% Furthermore, dividing the tokens into separate groups makes decoding easier for our simple one-layer decoder.
%\paragraph{Text Decoder}

We employ a simple shallow Transformer decoder to reconstruct the original input conditioned on the text groups. The shallow decoder has to rely on the representation in the groups for decoding, thus putting the burden on the encoder to better encode the information about the input in the groups \citep{bowman2015generating}. 

The output of this layer is 
% \vspace{-1ex}
\begin{equation}
    \overline{T}_i = \text{TransformerDecoder}(T_i|\{g^T_{i1} \dots g^T_{iK}\}).
% \vspace{-1ex}
\end{equation}
The probabilities from these predictions are then used to define the reconstruction loss:
% \vspace{-0.5ex}
\begin{equation}
    L_\text{reconstruction} = \sum_{i=1}^B \text{CE}(\overline{T}_i, T_i|\{g^T_{i1},\dots,g^T_{iK})\})
% \vspace{-0.5ex}
\end{equation}
where $\text{CE}$ is the cross entropy between the output probabilities of the decoder and the original input given the discovered groups. 
% This reconstruction loss from a simple decoder encourages the model to learn representations which are easy to map back to the original text.

\section{Related Work}
% \begin{itemize}
%     \item GroupViT
%     \item Object detection DETR, MDERT, DQ-DETR
%     \item Object discovery slot attention and all the relevant work here
%     \item open-vocabulary semantic segmentation
%     \item vision-language models with token-patch alignment like FILIP, 
%     \item phrase grounding 
%     \item referring image segmentation, shatter and gather \citep{kim2023shatter}
%     \item (meaningful) unit discovery in language: CANINE \citep{clark-etal-2022-canine}, Charformer \citep{tay2022charformer}, Dynamic pooling \citep{nawrot-etal-2023-efficient},  SA \citep{behjati2023inducing},  NVIB \citep{behjati-etal-2023-learning}
% \end{itemize}
Our work is related to different tasks in vision and language, which we will explain in this section. 

\paragraph{Object discovery.} Here the task is to discover the objects in an image or video without any supervision. Slot-based object discovery \citep{locatello2020object} has become popular due to the simplicity of the method \citep{singh2022illiterate,sajjadi2022object,singh2022neural_binder,seitzer2023bridging,singh2023neural,wu2023inverted_attention,wu2024structured_world_modeling, didolkar2025ctrl}. We have a novel adaptation of this method in discovering units similar to phrases in language with visually grounded semantics. 
% \citet{wu2023inverted_attention} demonstrated that by inverting the attention mechanism within a transformer model, it can be repurposed for object discovery. Others proposed binding mechanisms \citep{singh2022neural_binder} and a semantic vector-quantized autoencoder \citep{wu2024structured_world_modeling} to enhance the discovery of objects in images. 

\paragraph{Weakly supervised visual grounding.} Visual grounding refers to the tasks where a phrase or expression is grounded in the image. In the weakly supervised setup, the only information used is the pairing of the image with its caption \citep{datta2019align2ground,gupta2020contrastive,wang-etal-2020-maf,chen-etal-2022-contrastive,he2024improved,kuang2025momentum}. % only the information that the image and its caption are paired is used. 
In referring expression comprehension and referring image segmentation, the model must identify a specific part of the image described in a single expression. \citet{kim2023shatter} addressed the task of referring image segmentation by employing a slot-based object discovery module and merging relevant slots by cross attending over them with the textual query to build the final segmentation. %TODO: add more references 
In this task, the phrases are predetermined and no discovery happens on the language side, which is what our model is designed to do.
% \citet{chen2022contrastive_with_em} %Contrastive Learning with Expectation-Maximization for Weakly Supervised Phrase Grounding

\paragraph{Vision language models with vision and language alignments.} 
While many large-scale vision language models have been developed, it has been shown that they fall short in understanding fine-grained details in the image. This is especially more pronounced in the dual-stream Vision Language Models (VLMs) like CLIP \citep{radford2021learning}, where the modalities interact only via a single-vector representation. Therefore, there has been efforts to align language and vision at the level of patches and tokens \citep{yao2021filip,wang2022multi,mukhoti2023open, jing2024fineclip, zhang2024longclip, xie2025fgclip}. \citet{zeng2021xvlm} use additional supervision from phrase grounding annotations to help the model learn the alignments.  
\citet{bica2024improving} aligns tokens and patch
embeddings at different levels of granularity simultaneously. 
\citet{li2022finegrained_aligned} learns the semantic alignment from the perspective of game-theoretic interactions.
% Despite these advances, most models still rely on predetermined image patches or a single textual embedding, leaving room for discovering more meaningful, object-level alignments.
%%other references: 

\paragraph{Object detection.} The objective of this task is to detect the object boundaries in an image with a supervised objective. 
%where the models are trained with supervision of coordinates of the boundaries as well as the object class.
%The models are often trained given the annotations of the object boundaries and the object class. 
Our work is related to query-based object detection, such as the approach in \citep{detr,kamath2021mdetr}, where, at decoding time, learnable object queries attend to the input features and encode an object. 
\citet{liu2023dq} extend this approach by proposing a dual query model, demonstrating that simultaneously learning phrases and their corresponding objects improves the module's groundable understanding.
%Recently, \citet{liu2023dq} suggested learning text phrases as well as image objects by proposing dual query DETR and showed that learning phrases and their referent objects at the same time improves the groundable understanding of their model. 
The main difference between our model and this line of work lies in the weakly supervised nature of our approach. 

\paragraph{Zero-shot open-vocabulary semantic segmentation.} 
Semantic segmentation is a well-established task in computer vision. Recently, with the rise of VLMs, these models have demonstrated promising zero-shot capabilities in the semantic segmentation task as well.
\citep{xu2022groupvit} propose a hierarchical grouping architecture that learns to group image regions without pixel-level annotations, relying solely on paired image and text data. 
%groupvit
\citet{patel2023simcon} expanded on image-text alignment, suggesting to not only align an image to the corresponding text but also to the text from visually similar samples. Additionally, \citet{mukhoti2023open} %open vocabulary sem seg with path aligned
propose aligning patch tokens from a vision encoder with the <cls> token from a text encoder to enhance the model's performance. 
% \citet{ranasinghe2023perceptual}

\paragraph{Unit discovery in language.}
% CANINE \citep{clark-etal-2022-canine}, Charformer \citep{tay2022charformer}, Dynamic pooling \citep{nawrot-etal-2023-efficient},  SA \citep{behjati2023inducing},  NVIB \citep{behjati-etal-2023-learning}
Lately, discovering language units as part of the model architecture has been explored. These models operate on top of characters, where the units are usually at the level of subwords or words. The purpose is to optimize model efficency \citep{dai2020funnel_transformer,hierarchical_transformer,nawrot-etal-2023-efficient,sun-etal-2023-characters} or to skip the tokenization step of preprocessing and develop an end-to-end model \citep{clark-etal-2022-canine,tay2022charformer,cao-2023-best,behjati2023inducing,behjati-etal-2023-learning}. %Our work is related in that it also discovers units but it differs in that these units are semantically grounded to vision. 
Our research aligns with these developments by also focusing on language unit discovery. However, it differs in that these units are semantically grounded to vision. %, setting it apart from the solely text-focused methodologies previously mentioned.

%There is also another line of work which tries to model language at a higher level and then decode into the lower level \citep{nawrot-etal-2023-efficient, cao-2023-best}. 

%%citations added but not sure where to put
% \citet{sharma2018conceptual_captions} %dataset for automatic image captioning

\section{Experiments}

In this section, we empirically evaluate our proposed model. First, we probe the fine-grained vision-language understanding of our proposed text encoder under two benchmarks in \Cref{sec:probe}.  We show the effectiveness of our model in finding meaningful units by visualizing the attention maps in \Cref{sec:attn}. We then evaluate the quality of the discovered segments quantitatively by their accordance with human-annotated groundable phrases in \Cref{sec:seg}.  Finally, we analyze the contributions of different aspects of our model with a series of ablation studies in \Cref{sec:ablation}.

\subsection{Experimental Setup}
% \begin{itemize}
%     \item datasets (GCC3M, GCC12M, YFCC14M, Flickr30k, MSCOCO, )
%     % \item parameters
%     \item baselines
%         \begin{itemize}
%             \item groupvit pretrained text encoder as the upperbound
%             \item groupvit text encoder trained from scratch in the same setup as ours
%             \item spaCy for noun phrases as in DQ-DETR
%             \item clustering the text features of groupvit 
            
%         \end{itemize}
% \end{itemize}
\paragraph{Datasets:}
We trained our models on the training split of GCC3M dataset which consists of around 3 million image-caption pairs collected from the web \citep{sharma2018conceptual_captions}. %The average caption length in this dataset is $10.5$ tokens.  
We explain the datasets we used for evaluation in their corresponding sections and provide additional details in the Appendix. 

\paragraph{Parameters: }
We first resize the images to 224$\times$224 and then divide them into patches of size 16$\times$16. 
%After linearly projecting the patches, we add learnable positional encodings to them and then feed them to the image encoder. 
The image encoder has 12 Transformer encoder layers with the hidden dimension of 384 and two grouping blocks at the 6th and 9th layers. The number of groups in the first block is 64 and 8 in the second block. We load the weights from the GroupViT released checkpoint\footnote{We take the checkpoint trained on GCC3M \citep{sharma2018conceptual_captions}, GCC12M \citep{changpinyo2021cc12m} and YFCC14M \citep{thomee2016yfcc100m} datasets.}  \citep{xu2022groupvit} and keep it frozen during training. 

For the text encoder, we have 6 Transformer encoder layers followed by a grouping block\footnote{Our preliminary experiments with two blocks did not lead to reasonable results.} and then another 3 Transformer encoder layers. Each self-attention layer has 4 heads. 
Since the grouping vectors are learnable, the number of them needs to be fixed during training, so we treat the number of groups $K$ as a hyperparameter and tune it for different datasets. %, as in \citep{xu2022groupvit,seitzer2023bridging,locatello2020object}. 
We experiment with $K={1,2,4,8,16}$ as the number of groups and report the performance and results of the model trained with 4 groups as it has the best performance, and study the effect of having different numbers of groups in our ablations (\Cref{tab:ablation}). The text decoder has only 1 Transformer decoder layer consisting of one self-attention and one cross attention layer, each with 1 attention head. We tie the weights between the token embeddings in the encoder and the decoder. Both the encoder and the decoder have a model dimension of 128. The linear projection heads map each modality's feature vector to 256 dimension. We fix the $\tau$ to $0.07$ in our contrastive losses and $\lambda$ equals to 1. 
We use Byte Pair Encodings \citep{sennrich-etal-2016-neural} as our tokenizer with a vocabulary size of around 50k tokens and the maximum number of tokens is set to $M=77$ following previous work \citep{radford2021learning,xu2022groupvit}. 
We train our models with a batch-size of 4096 for 25 epochs and use the GradeCache library \citep{gao2021gradcache} to obtain this batch size on a single RTX3090 GPU\footnote{It takes around 48 GPU hours for every model to train.}. 
We trained our models with AdamW optimizer~\citep{loshchilov2017decoupled} with a learning rate of $0.0016$ with linear warmup for 2 epochs and cosine annealing decay.

\paragraph{Baselines:}
% We compared our model against a dual-stream vision language model where the text encoder has 9 Transformer layers, where the final text representation is taken from the <eos> token. This is the architecture used in GroupViT and other dual-stream vision-language models \citep{radford2021learning} and has approximately the same number of parameters as our proposed model. We train this model under the same training setup as our own model. 
% As for a fair comparison, we compare our models against a vision-language model with a Vanilla Transformer text encoder that has approximately the same number of parameters as our proposed model (i.e., 9 Transformer encoder layers) and train it under the same training setup. That is, we freeze the vision-encoder and only train the  text-encoder with the same training objectives as our models. We take the <eos> token as the final text representation.  This is the architecture used in GroupViT and other dual-stream vision-language models \citep{radford2021learning}. 

In order to showcase the effect of discovering groups as part of the architecture in vision and language models, we consider the following baseline. We replace our proposed Text Group Transformer with a vanilla Transformer encoder that has 9 Transformer encoder layers, thereby having approximately the same number of parameters as our model (referred to as \textit{baseline} in the experiments). This text encoder architecture is the one used in GroupViT and other dual-stream vision-language models \citep{radford2021learning}.  We train it under the same training setup as our own model to provide a fair comparison.  That is, we freeze the image-encoder and only train the text side from scratch with the same training data and objectives as our models. We take the final text representation from the encoded <eos> token. %This will provide a fair baseline for showcasing the effect of discovering groups as part of the architecture in the fine-grained vision and language understanding of our model. 

In addition, we report the results of the trained GroupViT model with its own text encoder (i.e., 12 Transformer encoder layers) and 2 layer projection heads (referred to as \textit{groupvit} in the experiments). Note that this model has many more parameters and has been trained on 10x more data.

\subsection{Fine-grained Vision-Language Understanding Probes}
\label{sec:probe}
% \citep{bugliarello-etal-2023-measuring} measured fine-grained understanding of VL models on four benchmarks: SVO, VSR, Valse and Winoground
We evaluate the fine-grained vision and language understanding of our model by employing different benchmarks which are specifically designed for this purpose. We will explain each of these benchmarks and the zero-shot performance of our models in the following sections. 
% \jamie{Is this the right place to say how the zero-shot classifier works?  Or repeat it in each section below.  My guess:}
In each case, the zero-shot classifier ranks the image-text pairs by their similarity scores $\text{sim}(\hat{z}^T_j,\hat{z}^I_i)$, which is the cosine between the pooled embeddings on the image and text sides. We refer to the score obtained from this zero-shot classifier as pair-wise ranking accuracy. 

\subsubsection{SVO Probes} 
\citet{hendricks-nematzadeh-2021-probing} designed a benchmark where they pair every sentence with two images, one positive and one negative. The negative images are selected in a controlled fashion where only either subject, verb or the object of the image is different from the original one. %The test split of this dataset contains around 30k examples. %We report the pairwise ranking accuracy results in this experiment. 
% \begin{itemize}
%     \item one caption with positive and negative images 
%     \item pairwise ranking results
% \end{itemize}

\begin{table}[t]
    \centering
    \begin{tabular}{@{}lrrrr@{}}
    \toprule
        Model & subj &verb&object&overall \\ \midrule
        random&50&50&50&50\\ \midrule
        \gray{groupvit} & \gray{81.6} & \gray{77.3} & \gray{91.7} & \gray{81.0} \\ 
        % transformer (contr+rec) & 80.1 & 69.0 & 87.6 & 74.7\\
        baseline&\textbf{80.5} & 69.5 & 89.0 & 75.3 \\
        % transformer (contr)& 82.3 & 74.2 & 90.9 & 79.0 \\
        % transformer (contr)& 80.0 & 73.6 & 89.0 & 77.8 \\
        % ours (4 groups) & 79.3 & 69.1 & 88.5 & 74.8 \\
        ours (4 groups)& 80.3 & \textbf{70.1} & \textbf{90.4} & \textbf{76.0}\\
        % ours (8 groups - 2lproj) & 81.4 & 68.2 & 89.5 & 74.8 \\

        % ours (contrstive) & 60.4 & 55.0 & 64.7 & 57.9 \\
        % ours (contrastive) &  78.2 & 72.6 & 89.1 & 76.9 \\
        % ours (rec) & 51.4 & 49.7 & 50.8 & 50.2 \\

    \bottomrule
    \end{tabular}
    \caption{The zero-shot pairwise ranking accuracy of different models on the test-split of SVO probes.}
    \label{tab:svo}
\end{table}
\Cref{tab:svo} shows the results of the zero-shot performance of different models under this benchmark. We observe that our model has a better overall performance compared to the Transformer baseline, which verifies our hypothesis that representing the language in a fine-grained and meaningful manner helps the fine-grained vision and language understanding of the model.  
The Transformer's single-vector representation succeeds in capturing information about subjects, but our multi-vector representation does a much better job of representing objects, and to a lesser extent verbs.
Both of these models are well above the random baseline.  The results for GroupViT's Transformer model are not comparable because it is trained on much more data, but we see that the resulting increase is much higher on verbs than on the the groundable phrases (subjects and objects) that our model is designed to represent as separate vectors.
% \melika{TODO: discuss the results}
% \jamie{I tried to add some more discussion}

\subsubsection{FOIL-COCO} 
\citet{shekhar-etal-2017-foil} propose FOIL-COCO dataset where for every image there is a correct caption and a "foil" one. The foil caption is different from the original caption by altering one of the nouns in the original caption into a foil one. %We evaluate the zero-shot performance of our model with pairwise ranking accuracy in \Cref{tab:foil} on the test split of this benchmark which has around $99$k examples. 
In \Cref{tab:foil}, we observe that our model demonstrates a remarkably good performance, outperforming the transformer model. This indicates that the noun understanding of our model has improved by learning fine-grained representations. Additionally, despite being trained on substantially less data than the GroupViT text encoder, our model performs nearly as well.
%achieving comparable performance. 
% findings
% test ~ 99k examples 
% \begin{itemize}
%     \item one image with true and false captions. (a specific noun has been changed in the false caption)
%     \item pairwise-ranking accuracy
% \end{itemize}
\begin{table}[t]
    \centering
    \begin{tabular}{lr}
    \toprule
         Model& accuracy  \\ \midrule
         random&50\\ \cmidrule{1-2}
         \textcolor{gray}{groupvit}& \textcolor{gray}{82.5}  \\
         % \textcolor{gray}{groupvit(all features)}&85.17 \\
         baseline &80.91\\
         % transformer (contrastive)&78.64\\
         ours (4 groups)&\textbf{81.68}\\
         % ours (8 groups - 2 layer proj)&83.10\\
         % ours (contrastive)&80.68\\
         % ours (contrastive)&78.66\\
         % ours(rec)&42.59\\
         
    \bottomrule
    \end{tabular}
    \caption{The zero-shot performance of different models on the test-split of FOIL-COCO benchmark.}
    \label{tab:foil}
\end{table}
% \subsubsection{Waterbirds}
% \textcolor{orange}{MB: we should probably remove this section as we cannot have any useful findings}

% robustness to spurious data correlations \citep{ranasinghe2023perceptual}
% \begin{itemize}
%     \item prompt: photo of a [waterbird|landbird].
%     \item pairwise-ranking accuracy 
% \end{itemize}
% \begin{table}[t]
%     \centering
%     \begin{tabular}{lrrr}
%     \toprule
%          ours& water&land&$\Delta$ \\ \midrule
%          waterbird&83.64&54.67& -28.97 \\
%          landbird&29.49&72.1&-42.61\\
%     \bottomrule
%     \end{tabular}
%     \caption{ours}
%     \label{tab:my_label}
% \end{table}
% % \begin{table}[t]
% %     \centering
% %     \begin{tabular}{lrrr}
% %     \toprule
% %          groupvit& water&land&$\Delta$\\ \midrule
% %          waterbird&42.83&7.94& -34.89 \\
% %          landbird&85.27&98.98&-13.70\\
% %     \bottomrule
% %     \end{tabular}
% %     \caption{groupvit}
% %     \label{tab:my_label}
% % \end{table}
% \begin{table}[ht]
%     \centering
%     \begin{tabular}{lrrr}
%     \toprule
%          ours& water&land&$\Delta$ \\ \midrule
%          waterbird&28.97&10.74& -18.22 \\
%          landbird&81.24&96.76&-15.52\\
%     \bottomrule
%     \end{tabular}
%     \caption{baseline transformer}
%     \label{tab:my_label}
% \end{table}
\subsection{Attention Visualization}
\label{sec:attn}

\begin{figure}[t]
    \centering
    \includegraphics[width=0.5\linewidth]{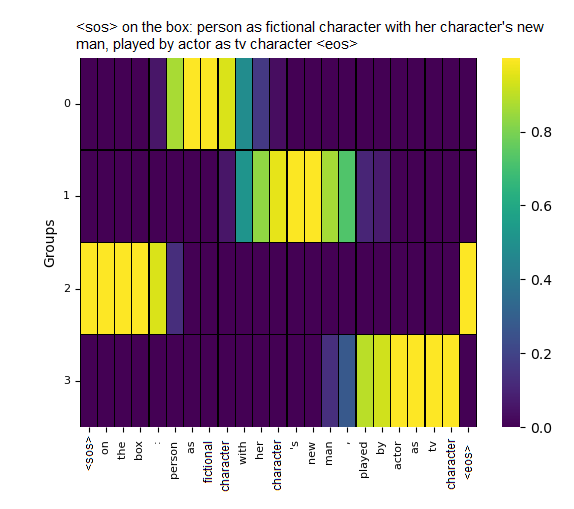}
    \caption{Soft attention of the groups over the input tokens. It shows that contiguous segments have emerged which capture phrase-like units.}
    \label{fig:attn}
\end{figure}

In order to understand what each group is representing, we visualize the soft attention weights of the groups over the input subwords in \Cref{fig:attn}. 
% \begin{itemize}
%     \item Contiguous segments emerging, explain why contiguous tokens are grouped together [highly correlated information in contiguous segments]
%     \item somewhat meaningful, which would be evaluated in the next section 
%     \item groups are at fixed positions, they seem to be starting as different positions in the input
% \end{itemize}
 Interestingly, we observe that contiguous segments have emerged, without imposing any contiguity constraints in the groupings. We believe that this is due to the fact that usually in language the contiguous tokens capture highly correlated information and that's why our model is grouping them together as part of its compression. Moreover, we can see that the emerging segments are meaningful in that they capture phrase-like units. We quantitatively evaluate the phrase discovery performance of our model in \Cref{sec:seg}. 
 %In \Cref{app:attn}, we visualize more attention maps. 
 %\melika{should we talk about how the groups look among different examples?
 %the groups look at a certain portion of the input, while the number of tokens each group binds to varies. 
 %}
 In our examination of a sample of attention maps, we observe that a given group tends to bind to similar positions in the text, but that the boundaries between groups vary.
 % \melika{TODO: In addition, we observe that groups are binding to different positions in the input and we validate this observation by averaging over all attention maps in a test set. We hypothesize that the initial group vectors are around certain input positions and the boundaries of every group are determined based on the input context. }
 
\subsection{Zero-shot Segmentation Evaluation}
\label{sec:seg}
% phrase discovery: segmentation quality in terms of IoU
% \begin{itemize}
%     \item Flickr30k validation set including 5K image-caption pairs
%     \item Hungarian matching algorithm with the objective of maximizing mIoU
% \end{itemize}

In order to evaluate the emerging segments in the attention maps quantitatively, we compare to human-annotated groundable phrases using several metrics (\Cref{tab:segmentation_eval}).
For the gold segmentation, we use the annotations in Flickr30k Entities \citep{plummer2015flickr30k} where groundable phrases are human-annotated. We report the results on the validation set of this dataset. % which has around 5000 examples. The number of annotated phrases in this dataset is on average $3.5$. 

We propose a metric similar to Intersection-over-Union (IoU) in the visual object detection literature which we call "tIoU". We first compute the soft attention weights of the groups over the input tokens. Then, by taking the argmax over the inputs, we have an assignment matrix of every input to a group. Given a gold segmentation, we can compute the IoU for each discovered group of tokens and each gold segment. For the computation of IoU, the intersection is equal to the number of overlapping tokens.  For the union, we do not count the tokens which were not annotated in the dataset, as the annotators did not have the constraint to include all the tokens in their annotation.  This gives us a matrix where by applying the Hungarian matching algorithm \citep{kuhn1955hungarian} maximizing this metric, we can obtain a 1-1 mapping between the discovered groupings and the gold segments. By having the mappings, we can compute precision, recall and F1 as well as IoU for each paired group and gold segment. In reporting the results, we first average every metric for the text input and then, report the average over all examples.  

% \begin{table}[t]
%     \centering
%     \begin{tabular}{@{}@{}lrrrrr@{}@{}}
%     \toprule
%          Model & mIoU&mic-IoU&P&R&F1\\ \midrule
%         random&30.68 &25.48&90.28&39.47&53.05 \\
%         % cluster-based&&&& &\\
%         rec-only &52.9/75.69&47.79&60.89&57.47&55.78 \\
%         ours&52.08/75.52&47.81&59.71&57.94&55.67\\
%     \bottomrule
%     \end{tabular}
%     \caption{Segmentation evaluation}
%     \label{tab:my_label}
% \end{table}
\begin{table}[t]
    \centering
    \begin{tabular}{@{}lrrrr@{}}
    \toprule
         Model & tIoU&P&R&F1\\ \midrule
        random&42.15&61.51&60.03&54.54 \\ \midrule
        %$M/4$&&&&&\\
        baseline (k-means)&52.77&61.82&64.87&59.55\\
        baseline (spectral-clustering)&38.88&49.81&52.82&45.52\\
        baseline (mean shift)&50.38&\textbf{99.64}&51.73&65.13\\
        % contr&&&&&\\
        % rec-only &75.69&76.18&87.35&85.15&83.41 \\
        ours (4 groups)&\textbf{76.42}&87.25&\textbf{85.83}&\textbf{83.72}\\
    \bottomrule
    \end{tabular}
    \caption{Phrase segmentation performance of different models under different evaluation metrics.}
    \label{tab:segmentation_eval}
\end{table}
In \Cref{tab:segmentation_eval}, we report the results of our evaluation. We compare our model against multiple baselines, including an untrained, randomly initialized text group transformer model. We also report the performance of applying different clustering methods over the encoded features of our transformer baseline. In particular, we apply k-means, spectral clustering \citep{shi2000spectral_clustering} and mean shift \citep{comaniciu2002mean_shift} with 4 clusters. %Finally, we report the performance of taking the mean ???
We observe that our model surpasses all the baselines by a large margin in almost all the metrics. Specifically, the high tIoU indicates that our model is indeed very good at discovering groundable phrases in the captions. 
% \melika{sell better, our model is really good here} 
% \melika{
% * implementing a cluster-based baseline (L=K-means)
% * compare with a stride-based baseline? not sure if this would be interesting
% } %TODO: implement a cluster-based baseline

\subsection{Ablation Study}
\label{sec:ablation}

In this section, we study the effect of different design choices (i.e., training losses and number of groups) on the performance of our models both in terms of groundable phrase discovery and fine-grained vision and language understanding. 

\begin{figure}[t]
    \centering
    \includegraphics[width=0.5\linewidth]{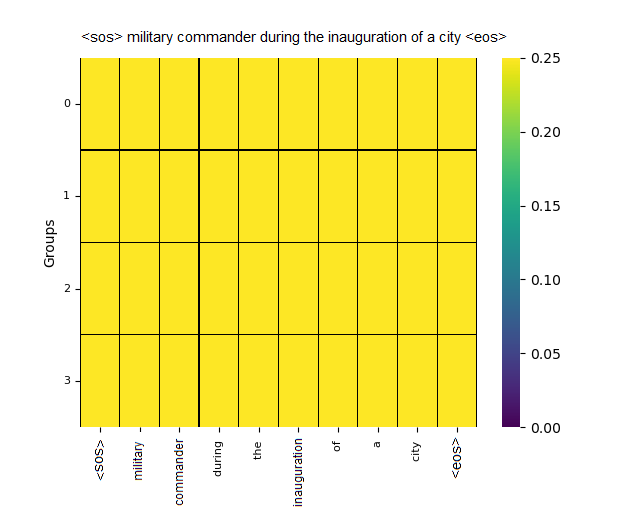}
    \caption{Soft attention of the groups over the input tokens for a model trained without the reconstruction loss. It shows a uniform attention map and lack of segmentation.}
    \label{fig:no-recon}
\end{figure}

\begin{table*}[t]
    \centering
    \begin{tabular}{lccccccc}
    \toprule
    & &\multicolumn{4}{c}{SVO}& \\ \cmidrule{3-6}
    model& tIoU& subject&verb&object&overall & FOIL-COCO& Noun Understanding\\ \midrule
    ours&\textbf{76.42}&\textbf{80.3}&70.1&\textbf{90.4}&76.0&\textbf{81.68}&\textbf{84.12}\\
    \hspace{0.5em}w/o contrastive loss&76.18&51.4&49.7&50.8&50.2&42.59&48.26\\
    \hspace{0.5em}w/o reconstruction loss&40.80&78.2&\textbf{72.6}&89.1&\textbf{76.9}&78.66&81.98\\
    % &&&&&&\\
    \bottomrule     
    \end{tabular}
    \caption{The performance of our model compared to the ablated ones on multiple datasets. Noun understanding refers to the average of performance on noun phrases (i.e. subjects, objects and FOIL-COCO). }
    \label{tab:ablation}
\end{table*}

\subsubsection{Training Losses}

In Table~\ref{tab:ablation} we see the different effects of the two types of loss on our model.  Without the contrastive loss, the model has no training on the image-text relationship, so it is not surprising that the image-text semantic evaluations are very low.  More surprisingly, although it still segments in a meaningful way without contrastive loss, the segmentation corresponds slightly less well to groundable phrases.  This suggests that semantic grounding in images actually helps the model discover meaningful units of text.

Interestingly, without the reconstruction loss, the model fails to segment in a meaningful way.  We can see this both in the tIoU score and in the uniform attention pattern shown in Figure~\ref{fig:no-recon}.  This lack of segmentation in turn affects the fine-grained understanding of the image-text relationship.  The holistic representations indicated by Figure~\ref{fig:no-recon} are relatively good at representing verbs, because verb understanding combines information across multiple objects.  But if we only consider the noun phrases (i.e.\ subjects, objects categories from SVO probes and FOIL-COCO), averaged in the last column, then segmenting the representation according to semantic objects, as indicated in \Cref{fig:attn}, results in much better understanding of the image-text relationship.
Therefore, both losses are crucial for the purpose of discovering groups of tokens with groundable semantics. 

% \begin{itemize}
%     \item effect of every loss term \melika{maybe move all the ablation results to a new table and then discuss}
%     \begin{itemize}
%     \item models trained with recon+contr have better noun understanding than the ones trained only with contr loss (foil it and also svo support this)
%     \item verb understanding [relation] has become worse. %why is that the case? 
% \end{itemize}
% \end{itemize}

\begin{table}[t]
    \centering
    \begin{tabular}{lrrr}
    \toprule
    \# of groups&tIoU&SVO&Foil\\ \midrule
         1&43.55&74.99&80.23\\
         2&53.12&75.30&80.01\\
         4&\textbf{76.42}&\textbf{76.0}&\textbf{81.68}\\
         8&63.93&74.8&80.56\\
         16&52.54&72.4&79.43\\
    \bottomrule
    \end{tabular}
    \caption{The performance of our model trained with different number of groups.}
    \label{tab:ngroups}
\end{table}

\subsubsection{Number of Groups}

% As mentioned before, we fix the number of groups as a hyperparameter in this work. 
In \Cref{tab:ngroups}, we investigate the effect of training our model with different numbers of groups. We can see that the model trained with 4 groups achieves the best results in all our evaluations. This implies that having too many or too few groups hurts the performance of our model and this hyperparameter should be tuned for different datasets and tasks.

\section{Conclusions}
% In conclusion, we developed a novel model for discovering units that are semantically aligned with the objects in the image. 

In this work, we developed a novel model for discovering meaningful units that are semantically aligned to the objects in the image. We freezed an image encoder which outputs groups that approximately represent objects and employ an analogous architecture on the text side to discover units that are at the level of phrases. While many dual-stream VLMs represent text as a single vector, we hypothesize that learning to represent language at a finer granularity will improve their fine-grained vision and language understanding. 

We verified our hypothesis by employing two specifically designed probing benchmarks, namely, SVO probes and FOIL COCO.
In addition, we showed that the segments that appear in the attention maps of groups attending to tokens are meaningful both qualitatively and quantitatively, in terms of overlap with human-annotated groundable phrases. Moreover, we ablated the effect of our losses on learning these units and concluded that both are necessary for having meaningful and semantically aligned units. 

Our experiments reveal the potential of our proposed model in specific real-world downstream applications such as semantic segmentation and concept binding, where fine-grained knowledge is essential. %We leave these investigations to future work.  

% \subsection{Downstream Evaluation Ideas}
% \begin{itemize}
%     \item Image-Text retrieval 1,1 
%     \item compositional image-text retrieval (WinoGround, negClip, vl text encoders are bottlenecks) 1,2
%     \item The dataset is not released. reconstruction ability (vl text encoders are bottlenecks, good reconstruction ability is a necessary but not sufficient condition) 1,2 
     
%     \item phrase grounding: zero-shot or fine-tune 3,1.5
%     \item semantic segmentation (COCO, Pascal VoC) 3,3
%     \item reverse phrase grounding
%     % \item probe  noun or verb understanding of the models (FOIL \citep{shekhar-etal-2017-foil}, SVO-probes\citep{hendricks-nematzadeh-2021-probing})
%     \item show the alignment by some means
%     \item evaluate some sort of robustness
%     \item interprete what every group has learned. we can look at the difference between the group vectors in the foil dataset (only one noun has changed)
% \end{itemize}

% \section{Discussion}
% \begin{itemize}
%     \item generalization to other datasets and tasks
%     \item downstream tasks application 
% \end{itemize}

\section*{Limitations}

% \begin{itemize}
%     \item language is only English
%     \item experiments at scale, training from scratch
%     \item (We don't learn the number of groups. ?)
    
% \end{itemize}
We have performed our experiments on the datasets and benchmarks in English. However, we do not make any language dependent assumptions in developing our model. Therefore, we believe that our method is generalizable across other languages as long as enough data for training is available.

We were not able to perform our experiments at scale due to the computational limitations. We expect that training the image and text encoder simultaneously from scratch would lead to better alignment between the two modalities, which should in turn improve our results. 

Lastly, our model by design is limited to a fixed number of groups, which is a common limitation in many object discovery works, such as \citet{xu2022groupvit,singh2022illiterate,seitzer2023bridging}. Therefore, for every dataset and task, it should be tuned as a hyperparameter for optimal performance. While this brings additional costs, it reduces the complexity of the problem we are tackling. Namely, the multi-modal setting is already very complicated and we need to make this simplifying assumption in order to make progress on the essential problem of finding groupings at all. 
Dynamically choosing the number of groups is a topic for future research.

% \section*{Acknowledgments}
% partly funded by NCCR

% \newpage
\bibliography{ anthology,custom,iclr2023_conference}
\bibliographystyle{tmlr}
\newpage
\appendix

% \section{Parameters}
% \label{app:param}
%Parameters For Packages

% \subsection{Statistics of data}
% GCC3M
% Foil: 99400 datapoints for test 
% SVO

% \section{More Attention Visualizations}
% \label{app:attn}

% \section{Model Parameters and Training}
\section{Details of the Datasets}
We provide additional details about the datasets we used in our work in the following. 
\begin{itemize}
    \item GCC3M: The average caption length in this dataset is 10.5 tokens.
    \item SVO probes: The test split of this dataset contains around 30k examples.
    \item FOIL-COCO: The test split of this 
benchmark which has around 99k examples. 
    \item Flickr30k Entities: The validation set of this
dataset has around 5000 examples. The number of annotated phrases in this dataset is on average $3.5$.
\end{itemize}
\section{Artifacts statements}
The datasets used do not have personally identifying information or offensive content. We provide the list of datasets used and the corresponding licenses in \Cref{tab:data_licenses}, which are all consistent with our academic use. 

\section{Descriptive Statistics}
Our results are from single runs for all the models trained.

\section{Packages}
We provide a list of packages used in our code in \Cref{tab:packages}.

\section{AI Assistants}
We utilized AI assistants for minor text editing and code completion tasks during the development of the model. 

% %Did you include information about your use of AI assistants?
% \section{Datasets}
% % Did you discuss the steps taken to check whether the data that was collected/used contains any information that names or uniquely identifies individual people or offensive content, and the steps taken to protect/anonymize it?
% %Data Contains Personally Identifying Info Or Offensive Content*
% The datasets used do not have personally identifying information Or offensive content.

% \subsection{Licenses}
% %Did you discuss if your use of existing artifact(s) was consistent with their intended use, provided that it was specified? 
% We provide the list of datasets used and the corresponding licenses in \Cref{tab:data_licenses}, which are all consistent with our academic use. 

\begin{table}[h]
    \centering
    \begin{tabular}{ll}
    \toprule
         Package & version   \\ \midrule
         Python & 3.7 \\ PyTorch & 1.8 \\
webdataset & 0.1.103 \\
mmsegmentation & 0.18.0 \\
timm & 0.4.12\\      
nltk & 3.8.1 \\ ftfy & 6.1.1 \\ regex & 2023.6.3 \\ 

    \bottomrule
    \end{tabular}
    \caption{The packages used in our code development}.
    \label{tab:packages}
\end{table}

\begin{table*}[tb]
    \centering
    \begin{tabular}{ll}
    \toprule
         Dataset& License  \\ \midrule
         GCC3M & Google license (\href{https://ai.google.com/research/ConceptualCaptions/termsAndConditions}{link})\\
         SVO-Probes & Creative Commons Attribution 4.0 International Public License (CC BY 4.0) \\
         FOIL-COCO & Creative Commons Attribution 4.0 License \\ 
         Flikr & Creative Commons Attribution 0: Public Domain \\
    \bottomrule
    \end{tabular}
    \caption{Datasets and their licenses.}
    \label{tab:data_licenses}
\end{table*}

\end{document}